\documentclass[11pt]{article}
\usepackage[preprint]{acl}
\usepackage{times,latexsym}
\usepackage[T1]{fontenc}
\usepackage[utf8]{inputenc}
\usepackage{microtype}
\usepackage{graphicx,booktabs,amsmath,amssymb,mathtools}
\usepackage{colortbl}
\usepackage{tabularx}
\usepackage{arydshln}
\usepackage{algorithm}
\usepackage{algorithmic}
\usepackage{tikz}
\usepackage{fontawesome5}
\usepackage{ragged2e}
\usepackage{needspace}
\usepackage{placeins}
\AddToHook{env/thebibliography/begin}{\RaggedRight\microtypesetup{expansion=false}}
\newcommand{\method}{VisAlloc}
\newcommand{\sg}{\operatorname{sg}}
\newcommand{\KL}{\operatorname{KL}}
\newcommand{\JSD}{\operatorname{JSD}}
\newcommand{\Q}{\operatorname{Q}}
\newcommand{\ind}{\mathbb{1}}
\hypersetup{colorlinks=true,citecolor=blue,linkcolor=blue,urlcolor=blue}
\title{VisAlloc: Allocating Visual Supervision for On-Policy Self-Distillation}
\newcommand{\authoremail}[1]{%
  {\normalfont\fontsize{8.5}{10.5}\selectfont\ttfamily\hypersetup{urlcolor=black}%
    \href{mailto:#1}{#1}}%
}
\author{
  \hypersetup{linkcolor=black}%
  \textbf{Haoran Xu\textsuperscript{1}\thanks{These authors contributed equally (co-first authors).},
    Tian Chen\textsuperscript{2}\footnotemark[1],
    Yuhe Wu\textsuperscript{3},
    Yanbo Zhao\textsuperscript{4},
    Junchi Liao\textsuperscript{5}} \\[1pt]
  \textbf{\hypersetup{linkcolor=black}Hongyu Wang\textsuperscript{6},
    Jiaze Li\textsuperscript{7},
    Wentao Wang\textsuperscript{8},
    Yanfeng Li\textsuperscript{9}\thanks{Corresponding authors.},
    Rui Qian\textsuperscript{2}\footnotemark[2]} \\[0pt]
  {\normalfont\fontsize{10}{12}\selectfont
    \textsuperscript{1}ZJU\hspace{0.75em}
    \textsuperscript{2}FDU\hspace{0.75em}
    \textsuperscript{3}HKUST(GZ)\hspace{0.75em}
    \textsuperscript{4}CUHK\hspace{0.75em}
    \textsuperscript{5}UESTC\hspace{0.75em}
    \textsuperscript{6}HNU\hspace{0.75em}
    \textsuperscript{7}Kuaishou Technology\hspace{0.75em}
    \textsuperscript{8}PKU\hspace{0.75em}
    \textsuperscript{9}MPU} \\[-1pt]
  {\renewcommand{\arraystretch}{0.8}%
    \begin{tabular}[t]{@{}l@{\hspace{1.5em}}l@{}}
      \authoremail{xhr964691257@163.com} &
      \authoremail{tianchen.fdu@gmail.com} \\
      \authoremail{p2424484@mpu.edu.mo} &
      \authoremail{qiianruii@gmail.com}
    \end{tabular}}
}
\hypersetup{pdfauthor={Haoran Xu, Tian Chen, Yuhe Wu, Yanbo Zhao, Junchi Liao, Hongyu Wang, Jiaze Li, Wentao Wang, Yanfeng Li, Rui Qian}}

\begin{document}
\maketitle
\raggedbottom
\renewcommand{\dbltopfraction}{0.95}
\begin{abstract}
On-policy self-distillation (OPSD) offers a promising approach to improving visual reasoning in multimodal large language models (MLLMs). However, not all samples benefit equally from privileged visual guidance, and uniform token supervision can dilute visually grounded signals. We propose \method{}, a dual-level framework combining sample-level visual filtering with token-level counterfactual reweighting. A fixed paired-teacher filter retains samples answered correctly from an evidence-centered positive crop but incorrectly from a spatially displaced negative crop. At the token level, both EMA teachers score the same student-generated response, and their absolute log-probability differences measure view sensitivity. Additional supervision targets tokens admitted by sensitivity or teacher-distribution disagreement, with shared frozen-prior, template-bigram, and student--teacher divergence gates. Inverse-frequency redistribution preserves the total additional weight within each micro-batch, while the positive teacher remains the sole distillation target. Inference requires only the full-image student. Across eight benchmarks, our 4B and 9B models achieve macro-averages of 81.44 and 84.92, outperforming Vision-OPD reproductions by 0.97 and 0.52 points, respectively. \href{https://github.com/Kaelyn01/VisAlloc}{\textcolor{black}{\faGithub}\hspace{0.25em}Code on GitHub}.

\end{abstract}

\begin{figure*}[t]
\centering
\includegraphics[width=\textwidth]{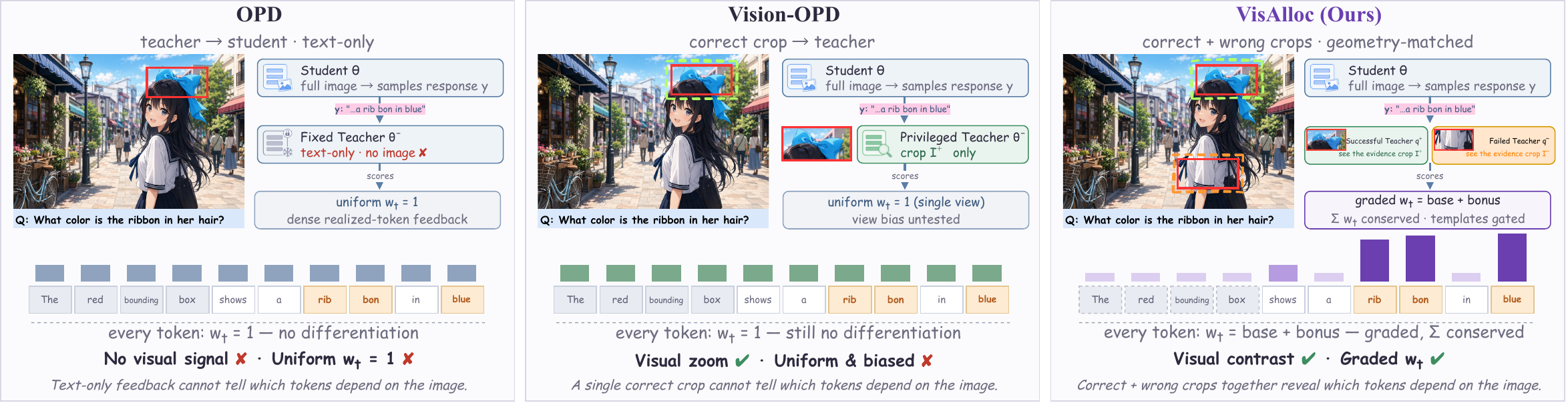}
\caption{Conceptual comparison of OPD, Vision-OPD, and \method{}. OPD provides uniform text-only teacher feedback. Vision-OPD adds a privileged evidence crop but retains uniform token weights. \method{} contrasts positive and negative teacher views to allocate a conserved token-weight budget. Fixed paired-teacher selection precedes allocation, and both teacher views are geometry-matched crops.}
\label{fig:teaser}
\end{figure*}

\section{Introduction}

On-policy self-distillation (OPSD) offers a promising approach to improving visual reasoning in multimodal large language models (MLLMs). Building on on-policy distillation, which supervises a student's own generated responses \citep{gkd,minillm}, OPSD derives teacher feedback from the model itself using privileged information \citep{opsd}. In the visual setting, Vision-OPD transfers privileged regional perception from a crop-conditioned teacher to a full-image student \citep{vision_opd}. This enables the student to learn from an evidence-centered view during training while requiring only the full image at inference.

However, effective use of privileged visual information requires considering both which samples receive supervision and how supervision is distributed across their response tokens. At the sample level, some examples can already be answered correctly without an evidence-centered crop, making them less compelling candidates for learning from privileged regional guidance. At the token level, relying exclusively on a single privileged teacher may yield biased supervision: its predictions reflect both visual evidence and linguistic priors, while its token probabilities alone do not reveal their dependence on question-relevant content. Consequently, distillation may emphasize confident textual continuations that remain similar even when access to the relevant evidence is reduced. These observations motivate a central question: \emph{how should privileged visual supervision be allocated across training samples and response tokens?}

Recent studies incorporate visual counterfactuals into distillation. OPD-V~\citep{opdv} selects tokens using a positive margin between zoomed and masked teacher views. CVPD discovers counterfactual visual blind spots \citep{cvpd}, and VAD reconstructs targets from visually attributable corrections \citep{vad}. Building on this line, we study the complementary roles of sample selection and token-level allocation in visual self-distillation. Our goal is to concentrate training on samples that remain challenging without privileged guidance and use contrasting visual evidence to refine supervision within their responses. Figure~\ref{fig:teaser} contrasts this design with uniform text-only and single-view visual distillation.

We propose \method{}, a dual-level optimization framework that combines \emph{sample-level visual filtering} with \emph{token-level counterfactual reweighting}. At the sample level, we exclude examples that the model can already answer correctly without access to an evidence-centered crop. This filtering focuses distillation on the remaining examples, for which privileged regional perception may provide useful guidance. Sample selection thus determines which examples enter distillation, while token reweighting determines how supervision is allocated within those examples.

To construct the counterfactual signal, we instantiate a \emph{successful teacher} and a \emph{failed teacher} from the same MLLM, with their roles defined by access to question-relevant visual evidence. The successful teacher observes an evidence-centered crop. Failed-teacher views are constructed using incorrect cropping to reduce access to the relevant evidence. The positive and negative views share the same rendering geometry to reduce variation unrelated to the intended visual intervention. These contrasting views provide a reference for assessing how teacher predictions change with the available visual information.

At the token level, both teachers score the same student-generated response under identical textual prefixes. Their absolute token log-probability differences measure view sensitivity, capturing changes in either direction and supplying information unavailable from the successful teacher's predictions alone. A continuous sensitivity weight modulates supervision across valid positions, while additional weight targets sensitivity or teacher-distribution disagreement under shared frozen-prior, template-bigram, and student--teacher divergence gates. An inverse-frequency rule then redistributes this additional weight while preserving its total within each micro-batch. Throughout training, the successful teacher remains the sole distillation target. Inference requires only the full-image student.

On eight benchmarks, \method{} achieves macro-averages of 81.44 at 4B and 84.92 at 9B, exceeding Vision-OPD reproductions by 0.97 and 0.52 points, respectively.

We make three contributions. First, a dual-level visual self-distillation framework that filters samples before applying counterfactual token reweighting. Second, a geometry-matched \mbox{positive--negative} crop construction, coupled with selective token weighting and inverse-frequency redistribution under a conserved micro-batch weight budget. Third, an eight-benchmark evaluation at 4B and 9B scales, demonstrating improvements over Vision-OPD reproductions at both model sizes.

\section{Related Work}
\paragraph{Text on-policy distillation.}
GKD~\citep{gkd} trains a student on its own generated sequences with teacher feedback, addressing the train--inference mismatch of distilling autoregressive language models, and admits alternative teacher--student divergences. MiniLLM~\citep{minillm} distills with reverse KL so that the student does not overestimate the low-probability regions of the teacher. The exposition by \citet{opd} makes the reward structure explicit: the student samples trajectories, and a fixed teacher grades every visited token through per-token reverse-KL feedback, combining the on-policy relevance of RL with dense supervision at a fraction of its cost. OPSD removes the external teacher: a single model acts as both, with its teacher copy conditioning on privileged information such as verified reasoning traces while the student copy sees only the question \citep{opsd}. ExOPD~\citep{exopd} recasts OPD as KL-constrained RL with a flexible reference model and a reward scaling factor, and extrapolating the reward beyond the teacher's scale lets the student surpass the teacher. This line relaxed the divergence, the teacher, and the reward scale in turn, while leaving the distribution of supervision across response positions largely uniform. SG-OPD~\citep{sgopd} combines phased sampling of verifier-approved teacher trajectories with token-level sign-consistency gating to address cold-start challenges and unreliable teacher guidance. CROP~\citep{crop} selects task-relevant response positions by comparing sensitivity to counterfactual text changes against sensitivity to meaning-preserving paraphrases.

\paragraph{Visual on-policy distillation.}
Vision-OPD~\citep{vision_opd} adapts on-policy distillation to fine-grained vision. Its teacher receives an evidence-centered crop, while its student sees the full image. This objective transfers visual zooming along student rollouts and requires no tool at inference. Video-OPD~\citep{videoopd} applies on-policy distillation to temporal video grounding. OPD-V~\citep{opdv} addresses modality imbalance, in which dominant textual signals crowd out visual input. It selects on-policy tokens whose modality-balance margin is positive. VA-OPD~\citep{va_opd} derives visual advantage from token-probability changes caused by fine-grained detail, reweights whole rollouts, and normalizes KL separately in two advantage groups.

\section{Method}
\label{sec:method}

\subsection{Preliminaries}
\label{sec:preliminaries}

We consider on-policy self-distillation for visual reasoning. Given a question $x_i$, a full image $I_i$, and a reference answer $a_i$, the student with parameters $\theta$ generates a response $y_i=(y_{i,1},\ldots,y_{i,T_i})$. The positive and negative teachers share the same model parameters but receive different visual evidence: the positive view $I_i^+$ is an evidence-centered crop, while the negative view $I_i^-$ reduces access to question-relevant evidence. During training, both teacher passes use the detached exponential-moving-average (EMA) state $\bar\theta$. The positive teacher supplies the distillation target, and the negative teacher provides a reference for measuring dependence on visual evidence. We denote the valid positions in a micro-batch by $\mathcal V$.

\paragraph{Shared trajectory and teacher state.}
The training full image contains a red target box, and the question instructs the model to focus on that box. The student first samples $y_i$ from the full-image rollout policy. For every fixed prefix $y_{i,<t}$, we then evaluate
\begin{align}
 p_{i,t} &= p_\theta(\cdot\mid I_i,x_i,y_{i,<t}),\\
 q_{i,t}^+ &= p_{\bar\theta}(\cdot\mid I_i^+,x_i,y_{i,<t}),\\
 q_{i,t}^- &= p_{\bar\theta}(\cdot\mid I_i^-,x_i,y_{i,<t}).
\end{align}
The two passes share the same detached EMA parameters $\bar\theta$, held fixed throughout the actor update. Matching the query, response, and prefix isolates the visual intervention from trajectory changes.

\paragraph{Framework overview.}
\method{} addresses two coupled allocation decisions. At the \emph{sample level}, we first construct paired positive and negative views, then apply a fixed filter based on the correctness difference between their teachers. At the \emph{token level}, both teachers score the retained samples' student-generated responses, and their contrast determines coefficients for the positive-teacher distillation loss. View construction and sample filtering are completed before optimization, while token coefficients are computed on on-policy trajectories during training. Figure~\ref{fig:method} summarizes the framework, and Algorithm~\ref{alg:batch} in Appendix~\ref{app:settings} describes an actor update.

\begin{figure*}[t]
\centering
\includegraphics[width=\textwidth]{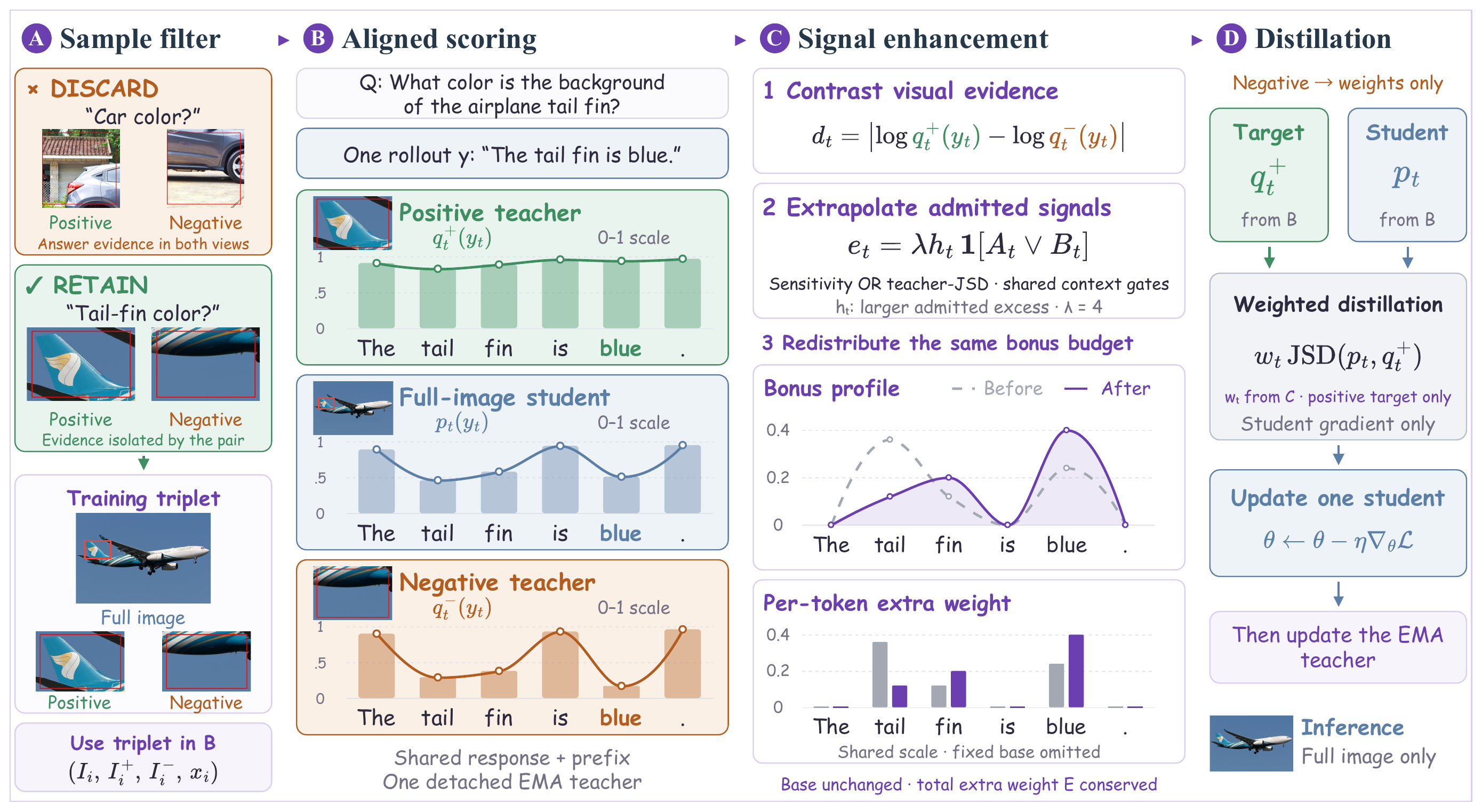}
\caption{Overview of \method{}. (A) Paired-view construction produces an evidence-centered positive crop and a spatially displaced negative crop. A fixed paired-teacher filter then retains samples that the positive teacher answers correctly and the negative teacher answers incorrectly, after annotation-quality checks. Each retained sample keeps its full image and paired teacher views. (B) Both teachers score the same student-generated response under aligned prefixes. (C) Realized-token sensitivity supplies the continuous base; sensitivity or teacher-distribution disagreement admits a bonus under shared context gates, and inverse-frequency redistribution preserves its total before mean normalization. (D) Only the positive teacher supplies the weighted JSD target, and EMA updates once after the actor update. The displayed redistribution is for $E>0$, and zero-budget batches retain zero bonus. The views and bonuses are schematic ($\lambda=4$, 4B).}
\label{fig:method}
\end{figure*}

\subsection{Sample-Level Signal Filtering}
\label{sec:sample}

\paragraph{Negative-view construction.}
Each candidate pairs an evidence-centered crop $I_i^+$ with one fixed negative crop $I_i^-$. We verify pixel-exact reconstruction of the official positive crop, then displace the target box to an intersection-over-union below $0.1$ with its original location. Both views preserve the crop dimensions, within-crop box geometry, 5-pixel frame, and $2\times$ LANCZOS resizing.

\paragraph{Sample filter.}
A fixed pre-training teacher answers each question under both views using identical prompts and decoding settings. Let $c_i^+$ and $c_i^-$ indicate whether these responses match the reference answer $a_i$. Among reliably annotated samples, we retain only positive-correct, negative-incorrect pairs:
\begin{equation}
 \mathcal D_{\mathrm{keep}}=\{i:c_i^+=1\ \land\ c_i^-=0\}.
\label{eq:sample-filter}
\end{equation}
Filtering runs once before optimization, and retained examples carry their full image and paired crops into training. 

\subsection{Token-Level Signal Enhancement}
\label{sec:gates}

For each retained sample, the paired teachers score the shared student-generated response defined in Section~\ref{sec:preliminaries}. Their realized-token contrast is
\begin{equation}
\begin{aligned}
 \Delta_{i,t}&=\log q_{i,t}^+(y_{i,t})-\log q_{i,t}^-(y_{i,t}),\\
 d_{i,t}&=|\Delta_{i,t}|.
\end{aligned}
\label{eq:contrast}
\end{equation}
For readability, later equations omit the sample index. The absolute contrast $d_t$ measures sensitivity in either direction, including cases where the evidence-centered crop lowers a sampled token's probability. It is a sensitivity signal rather than a correctness label, and the weighting rules below determine its effect on each token.

A continuous base and a gated bonus account for visual sensitivity while excluding recurrent templates from extra supervision. Inverse-frequency redistribution reallocates this bonus under a fixed budget.

\paragraph{Continuous base coefficient.}
Over the valid positions $\mathcal V$, the default configuration computes a dynamic knee and saturated sensitivity,
\begin{align}
 \tau_d &= \max\!\left(\Q_{0.9}(\{d_j:j\in\mathcal V\}),0.25\right),\\
 s_t &= \frac{d_t}{d_t+\tau_d},\qquad b_t=1+\gamma s_t,
\label{eq:base}
\end{align}
where $\Q_a$ is the empirical $a$-quantile and $\gamma=1$. Padding and positions without an available negative view are excluded from the quantile. The coefficient $b_t$ varies smoothly with sensitivity and preserves a nonzero distillation path for every valid position.

\paragraph{Context-gated bonus.}
Additional weight is admitted by realized-token sensitivity above the valid-token p80 ($\theta_d$) or teacher-distribution disagreement. Both branches require frozen rate $r(y_t)\geq0.12$, a non-template bigram, and detached student--positive-teacher JSD above the valid-token median $\theta_J$.

For the default mixing coefficient $\alpha=0.5$, let $m_t=(1-\alpha)p_t+\alpha q_t^+$ and define
\begin{equation}
 J_t=(1-\alpha)\KL(p_t\|m_t)+\alpha\KL(q_t^+\|m_t).
\label{eq:jsd}
\end{equation}
The implementation evaluates Equation~\ref{eq:jsd} on the student's top 100 vocabulary entries plus one residual probability bucket. Student and teacher use the same support, and all omitted mass is collected in that bucket. Thus $J_t$ is a coarsened distribution divergence rather than the full-vocabulary JSD. The detached gate and differentiable distillation loss use the same construction.

On the same support, $H_t$ is the positive--negative teacher JSD and $u_t=H_t/(H_t+\theta_H)$, where $\theta_H$ is its valid-token p90 knee, floored at $10^{-6}$ (Appendix~\ref{app:settings}). Writing $C_t=1$ when the local context is not in $\mathcal T$, the shared mask and bonus gates are
\begin{align}
 G_t={}&\ind[t\in\mathcal V]\ind[r(y_t)\geq0.12]\nonumber\\
      &\quad\cdot C_t\ind[J_t>\theta_J],\\
 A_t={}&G_t\ind[s_t>\theta_d],\nonumber\\
 B_t={}&G_t\ind[H_t>\theta_H],\\
 e_t={}&\lambda\max\!\left((s_t-\theta_d)_+,(u_t-0.5)_+B_t\right)\nonumber\\
      &\quad\cdot\ind[A_t\lor B_t].
\label{eq:bonus}
\end{align}
The preceding token contributes to $C_t$ only when valid and in the same response. Unseen types receive $0.10$ and fail the rate gate, and the first valid token has no preceding pair. The shared JSD gate avoids bonus when the positive-teacher loss is small, without guaranteeing accuracy gains.

\paragraph{Frozen selection prior and template set.}
Before optimization, one full-image response per training example is rescored under all three visual conditions. Within groups of 16 responses, we multiply saturated view sensitivity by the saturated absolute student--positive-teacher realized-token gap. Each factor uses its own 90th-percentile knee, floored at $0.25$ as in Equation~\ref{eq:base}. Positions strictly above the product's 90th percentile are selected. For token type $v$, occurrence count $n_v$ and selected count $k_v$ define
\begin{equation}
 r(v)=\frac{k_v+50\times0.10}{n_v+50}.
\label{eq:prior}
\end{equation}
Stored rates are rounded to four decimals. The template set contains adjacent pairs appearing in at least 25\% of responses, counted once per response. Both artifacts are frozen from training data. The prior uses a dual signal although the online base uses view sensitivity alone.

\paragraph{Frequency-aware redistribution under a fixed budget.}
Frequent eligible types may receive a large aggregate bonus simply because they recur. Let $E=\sum_{j\in\mathcal V}e_j$ and $a_t=(n_{y_t}+50)^{-1/2}$. We redistribute only the bonus:
\begin{equation}
 \widetilde e_t=
 \begin{cases}
 E\displaystyle\frac{e_ta_t}{\sum_{j\in\mathcal V}e_ja_j},& E>0,\\
 0,& E=0.
 \end{cases}
\label{eq:redistribution}
\end{equation}
For $E>0$, finite nonnegative counts make every $a_t$ positive and the denominator nonzero. Redistribution preserves zero bonuses and the base coefficient while assigning a larger share to less frequent types with the same initial bonus. It also preserves the exact total, $\sum_{t\in\mathcal V}\widetilde e_t=E$. The final coefficient is
\begin{equation}
 w_t=\frac{b_t+\widetilde e_t}{|\mathcal V|^{-1}\sum_{j\in\mathcal V}(b_j+\widetilde e_j)}.
\label{eq:normalization}
\end{equation}
Hence $\sum_{t\in\mathcal V}w_t=|\mathcal V|$, while positions without a negative view retain $w_t=1$. These are coefficient-sum identities: they do not preserve weighted loss, gradient norm, or the allocation after other importance weights. Redistribution can also concentrate the bonus on a small number of positions, and it gives no batch-independent upper bound on $w_t$.

\setlength{\dbltextfloatsep}{12pt plus 2pt minus 2pt}
\begin{table*}[!t]
\centering
\caption{Accuracy (\%) on eight benchmarks, grouped into fine-grained visual tasks and holdout tasks. All methods are evaluated using the same protocol, following Vision-OPD. $^\ddagger$\,Our reproductions of Vision-OPD and OPD-V. Within each model size, \textbf{bold} denotes the best score and \underline{underlining} the second-best, including ties.}
\label{tab:main}
\small
\setlength{\tabcolsep}{2pt}
\renewcommand{\arraystretch}{1.10}
\newcommand{\mainresulthead}[1]{\makebox[\linewidth][c]{\scriptsize\bfseries #1}}
\begin{tabular}{@{}>{\raggedright\arraybackslash}p{0.265\textwidth}>{\centering\arraybackslash}p{0.04\textwidth}*{9}{>{\centering\arraybackslash}p{\dimexpr(0.695\textwidth-20\tabcolsep)/9\relax}}@{}}
\toprule
 & & \multicolumn{4}{c}{\textbf{Fine-Grained Visual Tasks}} & \multicolumn{4}{c}{\textbf{Holdout Tasks}} & \\
\cmidrule(lr){3-6}\cmidrule(lr){7-10}
\raisebox{9pt}[0pt][0pt]{\footnotesize\textbf{Method}} & \mainresulthead{Size} & \mainresulthead{V*Bench} & \mainresulthead{Zoom} & \mainresulthead{HR-4K} & \mainresulthead{HR-8K} & \mainresulthead{MMVP} & \mainresulthead{CV-Bench} & \mainresulthead{MMStar} & \mainresulthead{POPE} & \mainresulthead{Avg.} \\
\midrule
\rowcolor{gray!10}\textit{Qwen3.5} & 4B & 84.29 & 47.69 & \textbf{84.38} & 80.13 & 76.67 & 87.13 & \underline{78.53} & 88.20 & 78.38 \\
SFT on Self-Teacher & 4B & 78.01 & 54.67 & 79.75 & 76.38 & 78.33 & 85.70 & 68.40 & 87.48 & 76.09 \\
GRPO~\citep{grpo} & 4B & 83.77 & 55.38 & 82.63 & 78.25 & \textbf{79.33} & 87.24 & 70.60 & 86.37 & 77.95 \\
DAPO~\citep{dapo} & 4B & 84.82 & 55.74 & 84.00 & 78.00 & \textbf{79.33} & 86.95 & 72.27 & 86.62 & 78.47 \\
OPSD~\citep{opsd} & 4B & 85.34 & 53.85 & 82.25 & 79.38 & \underline{78.68} & \underline{87.27} & 75.07 & 88.80 & 78.83 \\
Vision-OPD$^\ddagger$~\citep{vision_opd} & 4B & 89.01 & 59.41 & 82.75 & \underline{80.75} & 78.00 & 86.95 & 77.60 & \textbf{89.27} & 80.47 \\
OPD-V$^\ddagger$~\citep{opdv} & 4B & \underline{90.58} & \textbf{61.54} & \underline{84.25} & 80.50 & 78.00 & 85.86 & 78.13 & 88.53 & \underline{80.92} \\
\rowcolor[HTML]{E2D4F0}\textbf{\method{}} & 4B & \textbf{92.67} & \underline{60.71} & 83.00 & \textbf{82.12} & 77.00 & \textbf{88.00} & \textbf{79.00} & \underline{89.03} & \textbf{81.44} \\
\midrule
\rowcolor{gray!10}\textit{Qwen3.5} & 9B & 82.72 & 52.07 & 85.75 & 80.63 & \textbf{83.33} & 88.29 & \textbf{83.07} & 88.88 & 80.59 \\
SFT on Self-Teacher & 9B & 82.20 & 58.46 & 83.50 & 80.25 & 80.67 & 87.97 & 73.33 & 87.47 & 79.23 \\
GRPO~\citep{grpo} & 9B & 85.34 & 57.51 & 86.88 & 83.25 & 81.67 & 87.78 & 73.40 & 87.78 & 80.45 \\
DAPO~\citep{dapo} & 9B & 88.48 & 55.62 & 86.25 & 84.13 & 79.33 & 87.30 & 75.67 & 87.54 & 80.54 \\
OPSD~\citep{opsd} & 9B & 89.53 & 57.51 & 84.00 & 81.25 & 80.67 & 87.45 & 79.13 & 87.48 & 80.88 \\
Vision-OPD$^\ddagger$~\citep{vision_opd} & 9B & 91.10 & \textbf{64.97} & \textbf{88.25} & \textbf{86.50} & \underline{83.00} & \underline{88.66} & \underline{82.87} & 89.86 & \underline{84.40} \\
OPD-V$^\ddagger$~\citep{opdv} & 9B & \underline{92.83} & 64.01 & 86.72 & 82.68 & 80.31 & 88.24 & 80.95 & \underline{91.12} & 83.36 \\
\rowcolor[HTML]{E2D4F0}\textbf{\method{}} & 9B & \textbf{95.42} & \underline{64.77} & \underline{87.32} & \underline{84.64} & 81.40 & \textbf{91.25} & 82.34 & \textbf{92.18} & \textbf{84.92} \\
\bottomrule
\end{tabular}
\end{table*}

\subsection{Combined Objective and Algorithm}
\label{sec:objective}

\paragraph{Training objective.}
The positive teacher remains the sole distillation target. The fixed sample filter determines the training set, while the selected negative view affects training only through the token coefficients $w_t$. For response mask $M_t$, we minimize
\begin{equation}
 \mathcal L(\theta)=\frac{\sum_t M_t\,\sg(w_t\rho_tc_t)\,J_t(\theta,\bar\theta)}{\sum_t M_t},
\label{eq:objective}
\end{equation}
where $\rho_t$ is the inherited current-policy to old-policy importance ratio capped at $2$, and $c_t$ is the inherited rollout correction when present. Trajectory sampling and weight construction are detached, and gradients pass through the student side of $J_t$. The EMA teacher is updated once per actor update at rate $0.05$. Quantiles and frequency redistribution are computed within each local micro-batch, so packing and device topology are part of the effective algorithm.

\paragraph{Division of labor and cost.}
The sample-level stage constructs paired views and then applies the fixed paired-teacher filter once before optimization. Each retained example carries one spatially displaced negative crop into training, and each update adds one detached negative-teacher scoring pass. The pre-training filtering passes are separate from this per-update cost. The token-level stage changes only the coefficients applied to the positive-teacher loss. Neither stage adds a second student rollout or an inference-time crop operation.

\section{Experiments}
\label{sec:eval}
Table~\ref{tab:main} reports eight-benchmark comparisons at both model sizes, and Table~\ref{tab:ablations} presents the completed component and negative-view ablations.

\begin{table*}[!t]
\centering
\caption{Comparison with SOTA MLLMs. We report accuracy (\%) for each model. Among open-source models (single forward pass), the best results are highlighted in \textbf{bold}, and the second-best are \underline{underlined}. \method{}-9B achieves the highest six-benchmark average in this group. Baseline results are from \citet{vision_opd}.}
\label{tab:context}
\small
\setlength{\tabcolsep}{5.5pt}
\renewcommand{\arraystretch}{1.10}
\resizebox{\textwidth}{!}{%
\begin{tabular}{lcccccccc}
\toprule
\textbf{Model} & \textbf{Size} & \textbf{V*Bench} & \textbf{ZoomBench} & \textbf{HR-4K} & \textbf{HR-8K} & \textbf{MME-RW} & \textbf{MME-RW} & \textbf{Avg.} \\
 & & & & & & \textbf{EN} & \textbf{CN} & \\
\midrule
\multicolumn{9}{c}{\textbf{\textit{``Thinking-with-Images'' Agentic Models}}} \\
\midrule
DeepEyes & 7B & 85.86 & 46.51 & 75.13 & 72.63 & 64.10 & 64.09 & 68.05 \\
Thyme & 7B & 82.20 & 45.09 & 77.00 & 72.00 & 64.80 & 64.59 & 67.61 \\
DeepEyesV2 & 7B & 81.68 & 44.97 & 77.88 & 73.75 & 64.90 & 65.07 & 68.04 \\
SenseNova-MARS & 8B & 92.15 & 47.81 & 83.13 & 78.38 & 67.90 & 68.90 & 73.05 \\
\midrule
\multicolumn{9}{c}{\textbf{\textit{Closed-Source Models (Single Forward Pass)}}} \\
\midrule
GPT-5.2 & -- & 79.06 & 50.89 & 81.12 & 78.38 & 72.60 & 68.80 & 71.81 \\
GPT-5.4 & -- & 76.96 & 52.66 & 84.00 & 77.88 & 74.20 & 70.93 & 72.77 \\
Gemini-3.1-Pro & -- & 87.96 & 61.18 & 89.63 & 86.88 & 76.53 & 73.31 & 79.25 \\
Gemini-3.5-Flash & -- & 89.01 & 61.42 & 89.12 & 86.62 & 75.31 & 73.97 & 79.24 \\
\midrule
\multicolumn{9}{c}{\textbf{\textit{Open-Source Models (Single Forward Pass)}}} \\
\midrule
Qwen3.5 & 4B & 84.29 & 47.69 & 84.38 & 80.13 & 63.86 & 63.70 & 70.68 \\
Qwen3.5 & 9B & 82.72 & 52.07 & 85.75 & 80.63 & 71.40 & 67.67 & 73.37 \\
MiMo-VL-RL & 7B & 83.25 & 45.68 & 73.50 & 69.38 & 62.73 & 55.89 & 65.07 \\
Qwen3-VL-Instruct & 8B & 84.82 & 42.96 & 79.63 & 75.25 & 63.19 & 64.61 & 68.41 \\
ZwZ & 8B & 87.96 & 56.69 & 83.63 & 81.75 & 66.57 & 68.09 & 74.12 \\
MiniCPM-V-4.5 & 9B & 70.68 & 42.60 & 69.63 & 61.50 & 62.65 & 61.64 & 61.45 \\
GLM-4.6V & 106B & 86.91 & 50.06 & 82.13 & 78.88 & 65.57 & 65.62 & 71.53 \\
Qwen3-VL-Instruct & 235B & 91.10 & 56.09 & 86.13 & 80.38 & 71.74 & 69.04 & 75.75 \\
Qwen3.5 & 397B & 87.96 & 57.16 & \textbf{89.38} & \textbf{85.50} & 74.82 & 69.82 & 77.44 \\
Kimi-K2.6 & 1T & 88.48 & 53.14 & 81.88 & 78.00 & 69.22 & 66.13 & 72.81 \\
\midrule
\rowcolor[HTML]{E2D4F0}\textbf{\method{}} & 4B & \underline{92.67} & \underline{60.71} & 83.00 & 82.12 & \underline{75.43} & \underline{71.44} & \underline{77.56} \\
\rowcolor[HTML]{E2D4F0}\textbf{\method{}} & 9B & \textbf{95.42} & \textbf{64.77} & \underline{87.32} & \underline{84.64} & \textbf{76.90} & \textbf{73.59} & \textbf{80.44} \\
\bottomrule
\end{tabular}%
}
\end{table*}

\subsection{Experimental Setup}
\label{sec:audit}

\paragraph{Models.}
Our primary backbone is Qwen3.5-4B~\citep{qwen35}. The student receives the full image, while the positive and negative teachers share EMA parameters and observe evidence-centered and displaced crops, respectively (Section~\ref{sec:method}). Qwen3.5-9B examines transfer within the same model family (Section~\ref{sec:scale}).

\begin{figure}[!t]
\centering
\includegraphics[width=0.95\columnwidth]{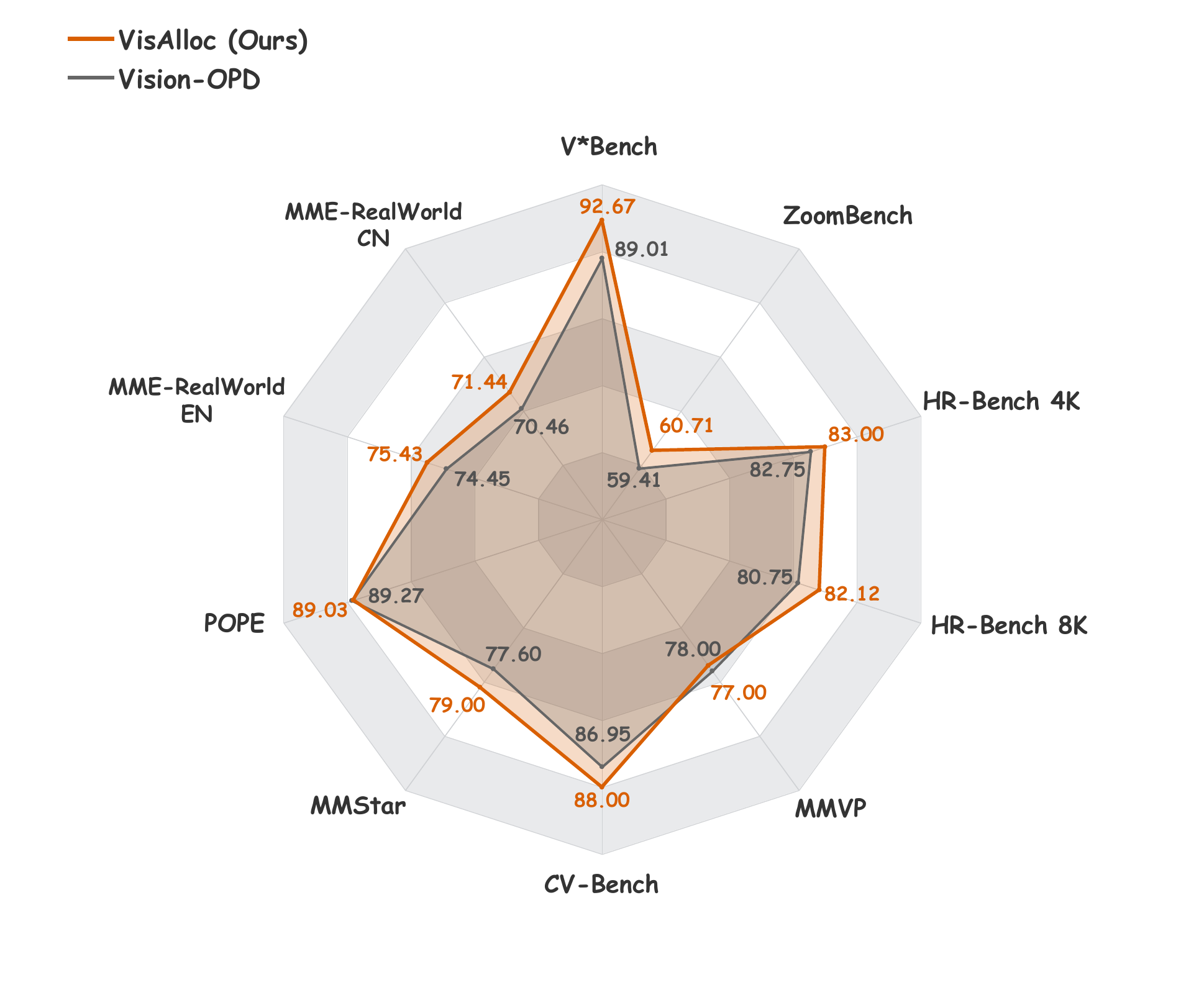}
\caption{Single-seed 4B results on ten benchmarks, with a shared 50--100\% radial scale. Orange/gray: \method{}/Vision-OPD. \method{} leads on eight axes and trails on MMVP and POPE. The eight-task averages (81.44/80.47) exclude MME-RealWorld EN/CN; polygon area is not an aggregate metric.}
\label{fig:benchmark-radar}
\end{figure}

\paragraph{Training data.}
We start from 6,241 paired visual-evidence examples derived from the Vision-OPD data. Each example carries a question, reference answer, full image, and geometry-matched positive and negative crops. The fixed pre-training filter retains 5,795 examples for which the positive teacher answers correctly and the negative teacher answers incorrectly, after annotation-quality checks (Section~\ref{sec:sample}). Training-signal diagnostics are reported in Appendix~\ref{app:frozen-records}.

\paragraph{Evaluation.}
We evaluate full-image inputs on eight benchmarks: V*Bench~\citep{vstar}, ZoomBench~\citep{zooming}, HR-Bench 4K/8K~\citep{hrbench}, MMVP~\citep{mmvp}, CV-Bench~\citep{cambrian}, MMStar~\citep{mmstar}, and POPE~\citep{pope}. We report their macro-average accuracy; the additional MME-RealWorld EN/CN results~\citep{mme_realworld} are reported separately. CV-Bench averages 2D and 3D accuracy, while MMVP uses question-level accuracy. Following Vision-OPD, evaluation disables thinking and uses greedy decoding with a 32,768-token limit. Scoring includes answer normalization, task-specific matching, and gpt-oss-120b judging. Appendix~\ref{app:protocol} provides the full scoring protocol.

\paragraph{Training and baselines.}
The training recipe uses 96 prompts per batch, 8 rollouts per prompt, a learning rate of $2\times10^{-6}$, 10 warmup steps, an EMA rate of $0.05$, and a 1,024-token response limit. We compare with the base model, self-teacher SFT, GRPO~\citep{grpo}, DAPO~\citep{dapo}, OPSD~\citep{opsd}, Vision-OPD~\citep{vision_opd}, and OPD-V~\citep{opdv}. All trained entries use seed 42, aligned data and update budgets, and the evaluation protocol described below. Appendix~\ref{app:settings} provides the implementation settings.

\subsection{Main Comparisons}
Table~\ref{tab:main} reports the eight-benchmark comparisons at both model sizes. In the 4B setting, \method{} achieves a macro average of 81.44, compared with 80.47 for our matched Vision-OPD reproduction, a gain of 0.97 points. At 9B, \method{} reaches 84.92 against 84.40, a gain of 0.52 points over our Vision-OPD reproduction under a single-seed protocol.

\begin{figure*}[!t]
\centering
\includegraphics[width=\textwidth]{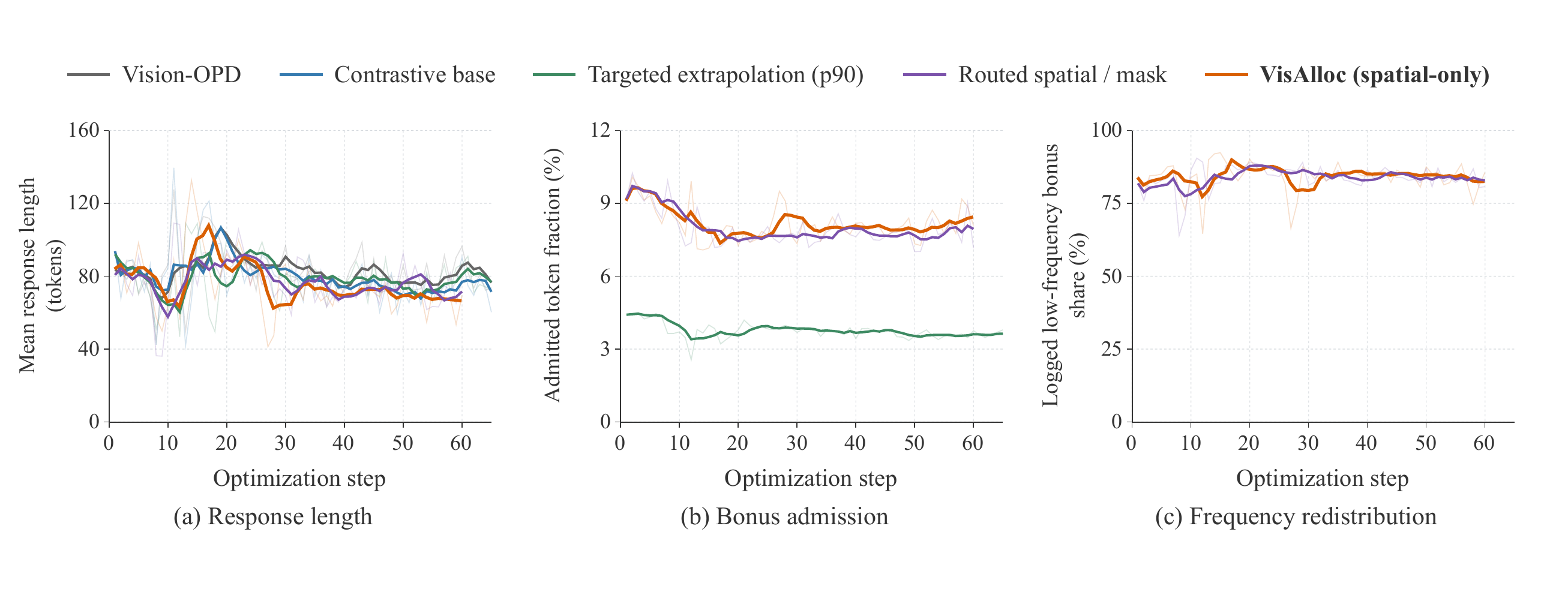}
\caption{Training dynamics of five 4B settings on a shared 0--65 step axis: (a) response length for five runs, (b) bonus admission for three, and (c) logged low-frequency bonus share for two. Orange: \method{} (spatial-only); gray/blue/green/purple: Vision-OPD, contrastive base, targeted extrapolation (p90), and routed spatial/mask. The first three runs extend to step 65; orange/purple end at step 60. (a) Mean response length decreases from $\approx$80--95 to 66--77 tokens; no collapse is observed over the recorded steps. (b) Final 10-step mean bonus admission is 8.2\% for \method{}, $\approx$2.3$\times$ p90 extrapolation (3.6\%) and slightly above routed spatial/mask (7.8\%). (c) Trailing 5-step means of the logged low-frequency share stay within 77--90\% (ending at 82.4/82.9; raw single-step values span 64--92\%). Thin traces show raw logs and thick traces show trailing 5-step means; the recorded budget ratio is 1.000.}
\label{fig:training-dynamics}
\end{figure*}

Figure~\ref{fig:benchmark-radar} compares \method{} and our Vision-OPD reproduction at 4B across the complete set of ten benchmarks. \method{} scores 75.43 versus 74.45 on MME-RealWorld EN and 71.44 versus 70.46 on CN, corresponding to a gain of 0.98 points on each of these two benchmarks. It leads on eight of the ten axes; the macro-average improvement over the matched Vision-OPD reproduction remains 0.97 points on the original eight-benchmark suite.

\paragraph{External-model comparison.}
Table~\ref{tab:context} places \method{} alongside the agentic, closed-source, and open-source models reported by \citet{vision_opd}. The baseline entries are reference scores reported in that work. The six columns cover V*Bench, ZoomBench, HR-Bench 4K/8K, and MME-RealWorld EN/CN; \method{} achieves average scores of 77.56 at 4B and 80.44 at 9B across this evaluation suite. The models in these groups differ in their scale, training data, and use of tools. The table therefore provides a broader context for Table~\ref{tab:main}, which compares methods within each model backbone.

\subsection{Transfer from 4B to 9B}
\label{sec:scale}
The 9B setting examines transfer within the same model family. Table~\ref{tab:main} includes 9B results for the base Qwen3.5 model, self-teacher SFT, GRPO, DAPO, OPSD, Vision-OPD, OPD-V, and \method{}. At this size, \method{} reaches a macro average of 84.92, against 84.40 for our Vision-OPD reproduction and 83.36 for OPD-V, gains of 0.52 and 1.56 points, respectively, under a single-seed protocol.

We construct the frozen responses, selection prior, template set, and frequency counts separately for the 4B and 9B backbones. Quantile-based gates depend on local micro-batch composition, so both settings use the same packing and parallelism configuration.

For each size $s\in\{4\mathrm B,9\mathrm B\}$, define the gain over Vision-OPD as $G_s=\operatorname{Acc}_s(\method)-\operatorname{Acc}_s(\text{Vision-OPD})$. Comparing these within-size gains separates method improvement from the larger backbone's absolute accuracy. Within this comparison, $G_{4\mathrm B}=0.97$ and $G_{9\mathrm B}=0.52$, so the higher absolute 9B accuracy should not be read as a larger method gain. A comparison of two model sizes would not, by itself, establish a scaling law.

\subsection{Ablations that Separate the Mechanisms}

Figure~\ref{fig:training-dynamics} reports response lengths for five selected 4B settings, bonus admission for three, and low-frequency bonus shares for two.

Table~\ref{tab:ablations} in Appendix~\ref{app:protocol} reports the sequential component ladder. On all 6,241 pairs, the contrastive base scores 80.12, targeted extrapolation with p90 admission reaches 80.84, and changing only the admission quantile to p80 reaches 81.12. Filtering retains 5,795 pairs and raises the macro average to 81.27; frequency redistribution reaches 81.35; and the teacher--teacher JSD OR branch completes \method{} at 81.44, a 0.97-point gain over the Vision-OPD reproduction. The lower block compares alternative negative-teacher views, with the spatially displaced crop attaining the highest score among the listed variants. Inference uses only the full-image student.

\FloatBarrier
\setlength{\dbltextfloatsep}{20pt plus 2pt minus 4pt}

\section{Conclusion}
\method{} combines fixed paired-teacher sample filtering with budget-preserving token allocation. From 6,241 source pairs, the filter retains 5,795 examples that are answered correctly from the evidence-centered positive crop and incorrectly from the spatially displaced negative crop. The negative teacher affects only token weights, while the positive teacher remains the sole distillation target, and inference uses only the full-image student. \method{} reaches macro averages of 81.44 at 4B and 84.92 at 9B, improving over the corresponding Vision-OPD reproductions by 0.97 and 0.52 points, respectively. These results support selective visual supervision while preserving the standard full-image inference procedure.

\clearpage
\section*{Limitations}
Our conclusions are limited to the evaluated Qwen3.5-4B and Qwen3.5-9B backbones and the reported benchmark suite. The macro average assigns equal weight to benchmarks of different sizes and difficulty, so it should be interpreted together with the per-benchmark results rather than as a universal measure of visual reasoning.

The fixed sample filter depends on teacher correctness under the paired crops and therefore inherits errors from answer matching and the teacher itself. A displaced crop can retain alternative evidence or make a question ambiguous even when its target box has little overlap with the positive view. Full-image training prompts also contain a target-box cue, so naturally phrased evaluation questions may differ from the training distribution.

The absolute realized-token contrast is model-internal and prefix-dependent. It can respond to irrelevant changes, and it is not a causal attribution estimate. A shared erroneous teacher can reinforce mistakes. The token prior and template filter depend on a tokenizer and prompting style. Rare valid terms can fail the prior gate, and common legitimate phrases can be filtered. The coarsened JSD omits distinctions inside its tail bucket. Micro-batch quantiles vary with packing, and inverse-frequency redistribution can concentrate weights. A fixed coefficient sum gives no guarantee of a fixed gradient scale or improved accuracy.

\section*{Ethical Considerations}
This work uses derived visual-question-answering data and inherits the licenses and potential biases of the underlying images and annotations. The code license does not override source-image usage terms. Increased visual sensitivity does not imply reliable perception in consequential applications. Emphasizing spurious visual attributes or underrepresented content unevenly may introduce additional performance disparities, and these effects are not measured here.

\bibliography{references}
\clearpage
\appendix
\flushbottom
\setlength{\textfloatsep}{12pt plus 2pt minus 2pt}

\section{Implementation and Reproduction Details}
\label{app:settings}
Table~\ref{tab:settings} lists the implementation settings used for the reported experiments. The 4B and 9B runs use backbone-specific frozen responses, priors, template sets, and frequency counts.

\begin{table}[ht]
\centering\small
\caption{Implementation settings for the reported 4B and 9B experiments. Backbone-dependent frozen artifacts are constructed separately for each model size.}
\label{tab:settings}
\begin{tabular}{@{}lr@{}}
\toprule
Setting & Value \\
\midrule
Backbones & Qwen3.5-4B / Qwen3.5-9B \\
Source / retained examples & 6,241 / 5,795 \\
Negative teacher view & Spatially displaced crop \\
Prompts / rollouts per prompt & 96 / 8 \\
Learning rate / warmup steps & $2\times10^{-6}$ / 10 \\
Epochs / GPUs per node & 1 / 4 \\
Prompt / response limit & 8,192 / 1,024 \\
EMA update rate & 0.05 \\
JSD mixing coefficient & 0.5 \\
Distillation support & Top 100 + tail \\
Distillation importance cap & 2.0 \\
Base knee quantile / floor & 0.9 / 0.25 \\
Base strength $\gamma$ & 1.0 \\
Bonus strength $\lambda$ & 4.0 (4B) / 2.0 (9B) \\
Sensitivity gate quantile & 0.8 \\
Student--teacher JSD quantile & 0.5 \\
Teacher--teacher JSD quantile & 0.9 \\
Frozen rate cutoff & 0.12 \\
Prior pseudo-count / base rate & 50 / 0.10 \\
Template frequency cutoff & 0.25 \\
Frequency pseudo-count & 50 \\
Frequency redistribution & Enabled \\
Stage-0 generation & Greedy; 96 tokens \\
Stage-0 prior grouping & 16 responses \\
\bottomrule
\end{tabular}
\end{table}

\paragraph{One actor update.}
Algorithm~\ref{alg:batch} updates the actor on fixed full-image student rollouts sampled after filtering and before optimization. Each mini-batch accumulates local micro-batch gradients before an optimizer step. Micro-batch losses from Equation~\ref{eq:objective} are backpropagated with inherited scale $\sigma_m$: the response-count ratio to the configured mini-batch size for dynamic batching, or the inverse accumulation count for fixed batching. The EMA teacher stays fixed across actor epochs and mini-batches, then updates once if the student was updated. Paired views use the same teacher state, and quantiles exclude padding.

\begingroup
\makeatletter
\let\visalloc@ruled\fs@ruled
\def\fs@ruled{\visalloc@ruled
  \def\@fs@mid{\kern2pt\hrule height0.6pt\kern2pt}%
  \def\@fs@post{\kern2pt\hrule height0.6pt\relax}}
\makeatother
\newcommand{\visallocmark}[1]{\tikz[remember picture,overlay]{\coordinate (#1);}}
\begin{algorithm}[t]
\small
\caption{\method{} Actor Update.}
\label{alg:batch}
\begin{algorithmic}[1]
\raggedright
\REQUIRE fixed full-image student rollout batch $\mathcal R$ from $\mathcal D_{\mathrm{keep}}$; frozen rate prior $r(\cdot)$; template set $\mathcal T$; gate quantiles $0.8/0.9$; bonus strength $\lambda$
\ENSURE updated student $\theta$ and EMA teacher $\bar\theta$
\FOR{each mini-batch $\mathcal B$ across actor epochs}
\STATE \visallocmark{va-outer-start}$\mathrm{ZeroGrad}(\theta)$
\FOR{each local micro-batch $\mathcal M$ in $\mathcal B$}
\STATE \visallocmark{va-inner-start}$y \gets \text{stored responses in }\mathcal M$; $K_t \gets \text{top-100 support at each position } t$
\STATE $q_t^{+}, q_t^{-} \gets \mathrm{Score}(y; I^{+}, I^{-}; \bar\theta)$ \COMMENT{detached EMA}
\STATE $d_t \gets \mathrm{Contrast}(q_t^{+}, q_t^{-})$ \COMMENT{realized token}
\STATE $\mathcal V \gets \mathrm{ValidPositions}(y)$; $s_t \gets \mathrm{Sensitivity}(d_t)$; $b_t \gets \mathrm{Base}(d_t)$; $J_t,H_t \gets \mathrm{Divergences}(y; \bar\theta)$
\STATE $\theta_d\gets\Q_{0.8}(s)$; $\theta_H\gets\max(\Q_{0.9}(H),10^{-6})$; $u_t\gets H_t/(H_t+\theta_H)$
\STATE $G_t \gets \mathrm{ContextGates}(r(\cdot),\mathcal T,J_t)$; $A_t\gets G_t\ind[s_t>\theta_d]$; $B_t\gets G_t\ind[H_t>\theta_H]$
\STATE $h_t\gets\max((s_t-\theta_d)_+,(u_t-0.5)_+B_t)$; $e_t\gets\lambda h_t\ind[A_t\lor B_t]$
\STATE $\hat e_t \gets \mathrm{Redistribute}(e_t)$ \COMMENT{Eq.~\ref{eq:redistribution}}
\STATE $w_t \gets \mathrm{Normalize}(\hat e_t)$ \COMMENT{Eq.~\ref{eq:normalization}}
\STATE $\mathcal L_m \gets \mathcal L(\theta;\mathcal M)$ \COMMENT{Eq.~\ref{eq:objective}}
\STATE \visallocmark{va-inner-end}$\mathrm{Backward}(\sigma_m\mathcal L_m)$ \COMMENT{accumulate gradients}
\ENDFOR
\STATE \visallocmark{va-outer-end}$\theta \gets \mathrm{OptimizerStep}(\theta)$ \COMMENT{after all micro-batches}
\ENDFOR
\IF{a student update occurred}
\STATE \visallocmark{va-ema}$\bar\theta \gets (1-\rho_{\mathrm{EMA}})\bar\theta + \rho_{\mathrm{EMA}}\theta$ \COMMENT{one EMA update}
\ENDIF
\RETURN updated student $\theta$ and EMA teacher $\bar\theta$%
\begin{tikzpicture}[remember picture,overlay,line width=0.35pt]
  \coordinate (va-outer-top) at ([xshift=-0.55\algorithmicindent,yshift=0.75\baselineskip]va-outer-start);
  \coordinate (va-outer-bottom) at ([yshift=-0.25\baselineskip]va-outer-end);
  \draw (va-outer-top) -- (va-outer-top |- va-outer-bottom);
  \coordinate (va-inner-top) at ([xshift=-0.55\algorithmicindent,yshift=0.75\baselineskip]va-inner-start);
  \coordinate (va-inner-bottom) at ([yshift=-0.25\baselineskip]va-inner-end);
  \draw (va-inner-top) -- (va-inner-top |- va-inner-bottom);
  \coordinate (va-ema-top) at ([yshift=0.75\baselineskip]va-ema);
  \coordinate (va-ema-bottom) at ([yshift=-0.25\baselineskip]va-ema);
  \draw (va-outer-top |- va-ema-top) -- (va-outer-top |- va-ema-bottom);
\end{tikzpicture}
\end{algorithmic}
\end{algorithm}
\endgroup

\paragraph{Distribution support.}
For the shared student-selected set $K_t$ of 100 vocabulary entries, define the coarsening map
\begin{equation}
 C_{K_t}(p)=\left((p(v))_{v\in K_t},\ 1-\sum_{v\in K_t}p(v)\right).
\end{equation}
The default divergence uses $C_{K_t}(p_t)$ and $C_{K_t}(q_t^+)$. Residual mass is computed stably in log space. Coarsening retains the probability outside $K_t$ but discards its internal vocabulary structure; top-$k$ selection is not differentiated.

The teacher-disagreement branch uses the same support and mixing coefficient:
\begin{equation}
\begin{aligned}
 H_t&=\JSD(C_{K_t}(q_t^+),C_{K_t}(q_t^-)),\\
 \theta_H&=\max(\Q_{0.9}(\{H_j:j\in\mathcal V\}),10^{-6}),\\
 u_t&=H_t/(H_t+\theta_H).
\end{aligned}
\label{eq:teacher-jsd}
\end{equation}

\paragraph{Frozen-data scope.}
Frozen scoring caps responses at 96 tokens; training permits up to 1,024 tokens. Each backbone has its own scoring archive, selection prior, template set, and frequency table, constructed from its bounded greedy responses.

\section{Sample Filtering and Negative-View Ablation}
The fixed pre-training filter applies the same teacher to evidence-centered positive and spatially displaced negative crops under identical prompts and decoding settings. After annotation-quality checks, it retains 5,795 of 6,241 pairs whose positive-view answers are correct and negative-view answers incorrect.

The main method uses spatially displaced negative crops for all retained samples. Mixed routing is restricted to this ablation: 4,116 samples use spatial crops and 1,679 use OPD-V-style masks. Its macro average is 81.16 versus 81.44 for spatial-only (Table~\ref{tab:ablations}).

\section{Reporting and Statistical Details}
\label{app:protocol}

\begin{table}[!t]
\centering
\caption{Sequential component and negative-view ablations. Upper-block rows cumulatively add the named component or change the admission quantile. Filtering is followed by frequency redistribution and then the teacher--teacher JSD OR branch. p90/p80 denote realized-token sensitivity admission. $\Delta$ is the preceding-row macro-average change (lower block: relative to the spatial-only row). All settings share the evaluation protocol and a 32{,}768-token generation limit. At step 60, the logged share of extra weight allocated to low-frequency token types is 85.6\% for the full recipe (routed: 80.7\%).}
\label{tab:ablations}
\footnotesize
\setlength{\tabcolsep}{3pt}
\begin{tabular}{@{}lrrr@{}}
\toprule
Training setting & Pairs & Avg. & $\Delta$ \\
\midrule
\multicolumn{4}{@{}l}{\textit{Sequential component ladder}} \\
Positive-only (Vision-OPD) & 6{,}241 & 80.47 & --- \\
\quad $+$ contrastive base & 6{,}241 & 80.12 & $-$0.35 \\
\quad $+$ targeted extrapolation (p90) & 6{,}241 & 80.84 & $+$0.72 \\
\quad $+$ p80 admission & 6{,}241 & 81.12 & $+$0.28 \\
\quad $+$ sample filtering & 5{,}795 & 81.27 & $+$0.15 \\
\quad $+$ frequency redistribution & 5{,}795 & 81.35 & $+$0.08 \\
\quad $+$ teacher-JSD OR (\method{}) & 5{,}795 & \textbf{81.44} & $+$0.09 \\
\midrule
\multicolumn{4}{@{}l}{\textit{Negative-view comparison} (full recipe)} \\
Text-only negative teacher & 5{,}795 & 77.85 & $-$3.59 \\
Routed spatial / mask & 5{,}795 & 81.16 & $-$0.28 \\
Spatial-only crop (\method{}) & 5{,}795 & \textbf{81.44} & --- \\
\bottomrule
\end{tabular}
\end{table}

\paragraph{Scoring protocol.}
All reported evaluations follow Vision-OPD: thinking is disabled, decoding is greedy, and generation is capped at 32,768 tokens. The scoring pipeline includes answer normalization, task-specific matching, and gpt-oss-120b judging.

The eight-benchmark main suite matches Vision-OPD and OPD-V (Table~\ref{tab:main}): fine-grained and high-resolution perception (V*Bench~\citep{vstar}, ZoomBench~\citep{zooming}, HR-Bench 4K/8K~\citep{hrbench}) and general multimodal capability (MMVP~\citep{mmvp}, MMStar~\citep{mmstar}, CV-Bench~\citep{cambrian}, POPE~\citep{pope}). Its macro average weights the eight accuracies equally. Additional MME-RealWorld EN/CN results~\citep{mme_realworld} are reported separately and excluded. CV-Bench averages 2D and 3D accuracy; MMVP uses question-level accuracy.

\section{Supplementary Signal Analysis}
\label{app:signals}

\subsection{Training-Signal Diagnostics}
\label{app:frozen-records}
All 5,795 retained spatial-only examples align one-to-one across the training Parquet, crop manifest, and three-view score archive. The archive contains 265,983 token positions and 9,189 types; its prior contains 9 template pairs and 934 types with rate at least $0.12$.

File hashes bind the manifest, training Parquet, token dump, and chat template to the same data and tokenizer. Prior and frequency files share the same tokenizer fingerprint. All retained examples record successful spatial-crop alignment and target-box IoU below $0.1$; 5,573 have zero target-box overlap.

\paragraph{Sensitivity is heterogeneous and signed.}
View-contrast magnitudes are long-tailed: median $0.0614$, 90th percentile $1.3427$, and 99th percentile $6.0189$ nats. Among recorded positions, 35.55\% have $\Delta_t<0$ and 14.06\% have zero recorded contrast. Rounded log-probability differences mean these zeros do not imply exact distributional equality. A negative contrast means the displaced view assigns greater probability to the realized token; it indicates neither correctness nor greater negative-teacher accuracy.

\paragraph{Repeated contexts require a contextual control.}
The pairs ``bounding box'' and ``red bounding'' occur in 54.39\% and 54.22\% of responses; ``inside the'' in 26.83\%. These frequencies reflect the red-box task format. Since ``red'' can express a valid visual attribute, globally removing the type would conflate formatting and content. The bigram filter therefore acts as contextual control rather than a global token blacklist.

\subsection{Frequency Concentration}

Statistics for the 4B training artifacts.

\paragraph{Concentration and eligibility.}
The 20 most frequent types cover 40.07\% of positions. Only 10.16\% of types meet the prior cutoff, but cover 25.55\% of positions. Type-level eligibility is not a uniform position-level retention rate; the other online gates still apply, so 25.55\% is not the bonus-eligibility rate. The 8,680 types occurring fewer than 50 times cover 16.35\% of positions. Rarity alone does not establish supervision value: frequency redistribution follows the contextual and JSD gates.

\end{document}